\documentclass[letterpaper, 10 pt, conference]{ieeeconf}  %

\IEEEoverridecommandlockouts                              %

\usepackage{graphics} %
\usepackage{epsfig} %
\usepackage{mathptmx} %
\usepackage{times} %
\usepackage{amsmath} %
\usepackage{amssymb}  %
\usepackage{amsfonts}
\usepackage[utf8]{inputenc}
\usepackage{siunitx}
\usepackage{hyperref}
\usepackage{subcaption}
\usepackage{forest}
\usepackage{tikz}
\usepackage[absolute,overlay]{textpos}
\usetikzlibrary{arrows.meta}

\definecolor{BLUE_ARROW}{RGB}{0,0,171}

\title{\LARGE \bf
TrueRMA: Learning Fast and Smooth Robot Trajectories with Recursive Midpoint Adaptations in Cartesian Space
}

\author{Jonas C. Kiemel$^{1}$, Pascal Meißner$^{1}$ and Torsten Kröger%
\thanks{$^{1}$Institute for Anthropomatics and Robotics – Intelligent Process Automation and Robotics (IAR-IPR),
	Karlsruhe Institute of Technology (KIT)
	{\tt\small \{jonas.kiemel, pascal.meissner\}@kit.edu}}%
}

\DeclareMathAlphabet{\mathcal}{OMS}{cmsy}{m}{n}

\tikzset{
	0 my edge/.style={densely dashed, my edge},
	my edge/.style={-{Stealth[]}, shorten >= 5pt},
	blue arrow/.style={-{Stealth[]}, draw=BLUE_ARROW, thick},
}

\newcommand{\myarrow}[1][0.25pt]{\tikz[baseline=-0.26em,y=3em, x=3em]{\draw[-{Stealth[]}, line width=#1] (0.0, 0.0) -- (0.6, 0);}}

\forestset{
	BDT/.style={
		for tree={
			if n children=0{s sep=35pt}{s sep=16pt},%
			thick,
			inner sep=0.75pt,    
			l=25pt,
			l sep=40pt, 
			edge={
				my edge,%
			},
			if n=1{
				child anchor=25,%
			}{child anchor=155},
			parent anchor=south,
		}
	},
	 arrow to/.style n args=1{%
	 	delay={%
	 		tikz+={%
	 			\draw [blue arrow] (#1) -- () ;
	 		},
	 	},
	 },
}
\begin{document}

\maketitle
\thispagestyle{empty}
\pagestyle{empty}

\begin{textblock*}{16.8cm}(2.5cm,0.5cm) %
	{\small © 2020 IEEE.  Personal use of this material is permitted.  Permission from IEEE must be obtained for all other uses, in any current or future media, including reprinting/republishing this material for advertising or promotional purposes, creating new collective works, for resale or redistribution to servers or lists, or reuse of any copyrighted component of this work in other works.}
\end{textblock*}

\begin{abstract}

We present TrueRMA, a data-efficient, model-free method to learn cost-optimized robot trajectories over a wide range of starting points and endpoints. 
The key idea is to calculate trajectory waypoints in Cartesian space by recursively predicting orthogonal adaptations relative to the midpoints of straight lines. We generate a differentiable path by adding circular blends around the waypoints, calculate the corresponding joint positions with an inverse kinematics solver and calculate a time-optimal parameterization considering velocity and acceleration limits.
During training, the trajectory is executed in a physics simulator and costs are assigned according to a user-specified cost function which is not required to be differentiable.   
Given a starting point and an endpoint as input, a neural network is trained to predict midpoint adaptations that minimize the cost of the resulting trajectory via reinforcement learning.
We successfully train a KUKA iiwa robot to keep a ball on a plate while moving between specified points and compare the performance of TrueRMA against two baselines. The results show that our method requires less training data to learn the task while generating shorter and faster trajectories.

\end{abstract}

\section{INTRODUCTION}

Generating a robot trajectory for a movement between a specified starting point and endpoint is typically divided into two stages \cite{kunz2012time}: 
The first step is finding waypoints that form a collision-free geometric path with a sampling-based motion planner, while timestamps are added to the waypoints in a second step. 
This approach assumes that suitable waypoints can be found without considering the timing of the trajectory execution. 
However, if the environment changes over time or as a consequence of the robot movement, the stages can no longer be treated as decoupled. 
Examples include collision-free path planning in the presence of moving obstacles \cite{baumann2001robot}, trying to transport a full drinking glass without spilling water or keeping a ball on a plate during the movement of the robot. 
While optimization-based planners \cite{zucker2013chomp, kalakrishnan2011stomp} can be used to generate a single trajectory between two specified points, we focus on the more challenging problem of learning cost-optimized movements over a wide range of starting points and endpoints.
Once trained, learning approaches can generate well-performing trajectories within a short amount of time, which is especially important if the endpoint is not known in advance. 
However, the gain in time comes along with a potentially time-consuming and data-intensive training phase. 

\begin{figure}[t]
	\centering
	\includegraphics[width=\linewidth, origin=c]{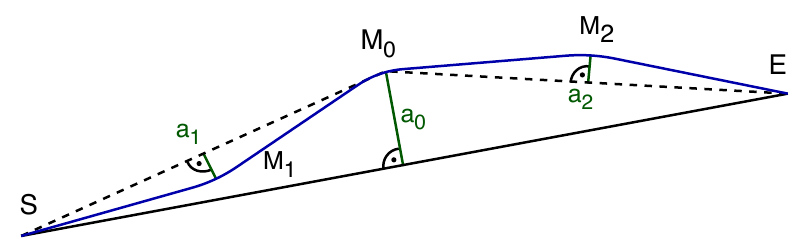}
	\caption{Trajectory generation with TrueRMA. 
	The final trajectory after two recursive adaptation steps is shown in blue. 
	The elements of an action vector $a_{0..2}$ define the length of the orthogonal adaptation vectors (shown in green).\newline
	$S$: starting point; $E$: endpoint; $M_{0..2}$: adapted midpoints}
	\label{fig:RMA}

\end{figure}

Learning arbitrary trajectories is hard due to the high-dimensional search space, a problem which is commonly known as the curse of dimensionality. 
Hence, the key to data-efficient trajectory learning lies in the reduction of the search space by incorporating knowledge of reasonable waypoint distributions and time-parameterizations. TrueRMA is based on the following assumptions:
\begin{itemize}
    \item The predicted path is preferred to be short. For this reason, TrueRMA starts with a straight line between the specified starting point and endpoint. 
    \item Waypoints should have similar distances and should be evenly spread between the starting point and the endpoint. As shown in Fig. \ref{fig:RMA}, this can be achieved by predicting orthogonal midpoint adaptations in a recursive manner. %
	\item With the aim of reducing cycle times, robot movements for industrial applications are preferred to be fast. %
    Hence, we parametrize all trajectories to be time-optimal with respect to the joint limits of the robot. %
\end{itemize}

Given that the assumptions are valid, TrueRMA provides an efficient way to reduce the dimensionality of the search space.
We demonstrate the efficiency of our approach by comparing the data required to learn a ball-on-plate task against two baselines: Direct prediction of waypoints and prediction of adaptations relative to the current end-effector pose.   
We note that the above-mentioned assumptions limit the scope of applicability to some degree.
TrueRMA works best if spreading the waypoints evenly between the starting point and endpoint is a reasonable strategy.
It is not directly applicable to tasks that require fast and slow segments within a trajectory or complex path shapes like loops.

\begin{figure*}[t]
	\centering
	\vspace{0.11cm}
	\includegraphics[width=\linewidth, origin=c]{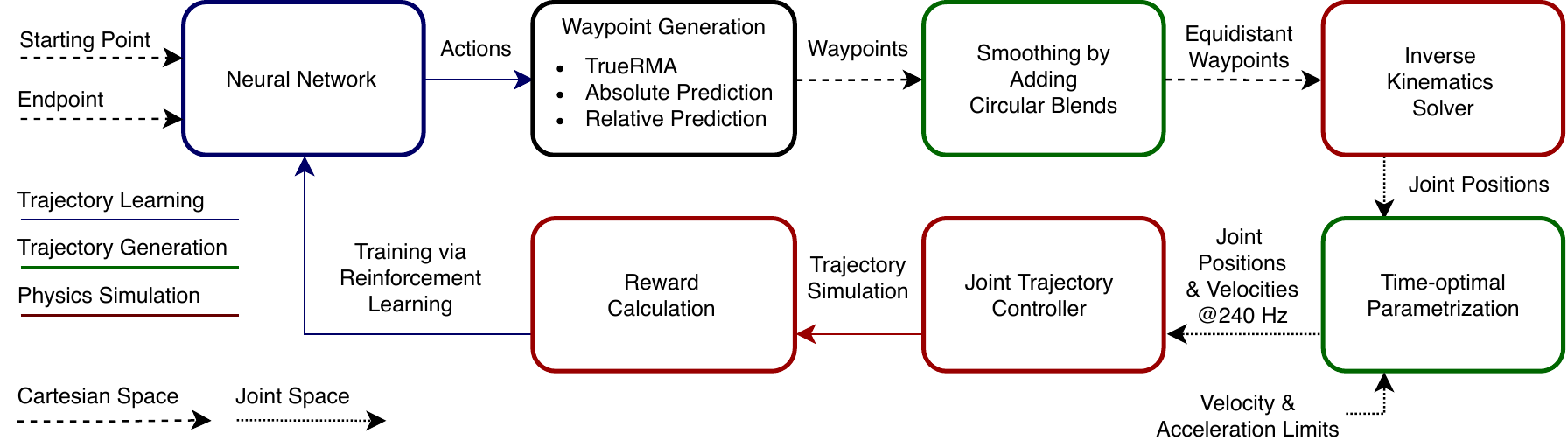}
	\caption{Overview of the system components}
	\label{fig:overview}
\end{figure*}

\section{Related Work}

\subsection{Motion Planners}
TrueRMA is related to offline motion planners, as the aim is to generate a trajectory specified by a starting point and endpoint.
However, traditional motion planners generate a path for a single trajectory, while TrueRMA learns trajectories over a wide range of starting points and endpoints, which allows fast waypoint generation at runtime. Given a list of waypoints, a differentiable path can be produced by adding circular blends \cite{kunz2012time, biagiotti2008trajectory}.
Examples for offline motion planners are sample-based methods like
probabilistic roadmap planners \cite{geraerts2004comparative} and optimization-based
algorithms.   In \cite{vallejo2000adaptive}, collision-free paths are generated based on recursive midpoint adaptations. Optimization-based motion planners include gradient-based techniques
like CHOMP \cite{zucker2013chomp} and gradient-free algorithms like STOMP \cite{kalakrishnan2011stomp}.
Gradient-based techniques require a differentiable cost function, while gradient-free approaches can handle arbitrary cost functions. 
TrueRMA is a gradient-free method. 
The STOMP algorithm is the closest to our approach.
The basic principle is to generate a large number of noisy trajectories based on an initial solution.
The cost function is evaluated for each noisy trajectory and an updated trajectory is generated by weighting each trajectory inversely proportional to its cost. The procedure is repeated in an iterative manner.
However, in contrast to our method, the generated trajectories are not time-optimal.
Instead, a fixed trajectory duration is chosen. Assuming a fixed execution time prevents the method from being applied for highly dynamic tasks, as the robot needs a certain dynamic reserve to execute potentially longer trajectories within the same time. While the STOMP algorithm works in joint space, TrueRMA predicts waypoints in Cartesian space.

\subsection{Reinforcement Learning in Robotics}
Deep reinforcement learning has been successfully applied to a wide range of robot skills like grasping \cite{berscheid2019robot},
locomotion \cite{tan2018sim} or in-hand manipulation \cite{DBLP:journals/corr/abs-1808-00177}.
Dynamic movement primitives (DMP) \cite{schaal2006dynamic} have been proposed to learn discrete and rhythmic movements.
DMPs can be trained with the $\mathrm{PI}^2$ algorithm \cite{theodorou2010learning}.
While DMPs generate smooth trajectories, they are not guaranteed to be time-optimal.

\section{System Overview}

The most important system components for trajectory learning with TrueRMA are shown in Fig. \ref{fig:overview}.
While the next paragraph aims to describe the workflow as a whole, further details on the individual components can be found in the following sections.

Given a Cartesian starting point and endpoint of a trajectory as input, a neural network predicts actions, which are used to generate waypoints in Cartesian space. We present three different strategies to generate these waypoints in chapter \ref{waypointGeneration}.
As explained in chapter \ref{trajectoryGeneration}, the path of a trajectory needs to be continuously differentiable. Simply connecting the generated waypoints with straight line segments leads to discontinuous joint velocities and acceleration peaks. After smoothing the path by adding circular blends around the waypoints, equidistant waypoints are sampled from the path. These waypoints are given to an inverse kinematics solver, which returns the corresponding joint positions.
Using a method proposed in \cite{kunz2012time}, we calculate a time-optimal trajectory parameterization and sample joint positions and velocities with a uniform time step in between points.
The trajectory is then executed by a physics simulator using a joint trajectory controller. 
During training, the performance of the trajectory is rated with a scalar reward, which is used to improve the network predictions via reinforcement learning. 
The computation times for the individual components are given in TABLE \ref{table_computationTime}.

\begin{table}[h]
	
	\centering
	\begin{tabular}{c c}
		\hline
		System component & Computation time \\
		\hline
		Neural network (action prediction) & $<$ \SI{1.0}{\milli\second} \\
		Waypoint generation (TrueRMA) & $<$ \SI{1.5}{\milli\second}\\
		Smoothing (circular blends) &  $<$ \SI{0.5}{\milli\second}\\
		Inverse kinematics solver &  $\sim$ \SI{50}{\milli\second}\\
		Time-optimal parameterization & \SI{100}{\milli\second} - \SI{260}{\milli\second} \\
		Trajectory simulation and reward calculation &  \SI{100}{\milli\second} - \SI{220}{\milli\second} \\
		\hline
	\end{tabular}
	\caption{Computation times for the three-dimensional ball-on-plate task using a single CPU core of an Intel i5-7300U}
	\label{table_computationTime}
\end{table}

\section{Trajectory Learning}
\subsection{Formalization}
We formalize trajectory learning as a one-step Markov decision process $(\mathcal{S}, \mathcal{A}, R_a)$, where $\mathcal{S}$ is the state space,  $\mathcal{A}$ is the action space and $R_a$ is the reward for action $a$.
Using model-free reinforcement learning, a policy $\pi: \mathcal{S} \mapsto \mathcal{A}$ is trained to map states $s \in \mathcal{S}$ to those actions $a \in \mathcal{A}$ that maximize the expected reward. The policy is represented by a fully-connected neural network with two hidden layers of size [200, 100].
The state is composed of a starting point and an endpoint in Cartesian space.
With TrueRMA, each waypoint is described by a fixed number of parameters. A single action specifies all waypoints of a path, which means that the dimensionality of the action space scales linearly with the number of predicted waypoints. 
Both state space and action space are continuous and normalized, meaning that each element of the corresponding vectors is in the range of [-1, 1]. 
Reward is calculated for each trajectory according to a task-specific performance metric.

\subsection{Implementation}
We use the OpenAI Gym toolkit \cite{1606.01540} to implement our environment and Ray \cite{DBLP:journals/corr/abs-1712-05889} for distributed training.  
The neural network is trained using Proximal Policy Optimization (PPO) \cite{schulman2017proximal}, an on-policy actor-critic method, which is known for its stability and reliability. 
The implementation of PPO is provided by RLlib \cite{liang2017rllib}. 
In order to make the comparison with the baselines as fair as possible, we did not tune hyperparameters.

\section{Physics Simulation}
TrueRMA allows to optimize robot trajectories based on their effects on the environment. 
We use PyBullet \cite{coumans2016pybullet}, a general-purpose physics engine, to simulate a KUKA iiwa robot and its surroundings. 
We note that the physics engine can by replaced by real-world data collection in future work. 

\subsection{Inverse Kinematics Solver}
Cartesian waypoints are converted to joint space by an inverse kinematics solver.
The waypoints are sampled equidistantly from the smoothed path.  
We use a numeric solver provided by PyBullet, which favors the solution that is the closest to a given rest pose.
By specifying a fixed rest pose, we ensure a unique and deterministic return value for each Cartesian waypoint despite the kinematic redundancy of the manipulator.      
Singularities are avoided by restricting the workspace accordingly.

\subsection{Joint Trajectory Controller}
The trajectory execution is performed by a constraint-based joint trajectory controller which accepts position and velocity setpoints. 
We run the physics simulator at a frequency of \SI{240}{\hertz}.
As the time-optimal trajectory is sampled at the same frequency, setpoints can be commanded at each simulation step. 

\subsection{Reward Calculation}
During training, a scalar reward that indicates the performance of the trajectory is calculated.
The full state of the environment during or after the trajectory execution can be considered to calculate the reward.
While TrueRMA does not require a differentiable cost function, learning is accelerated by providing a dense performance metric.

\section{Waypoint generation}
\label{waypointGeneration}
\subsection{Generating Waypoints in Cartesian Space}
Trajectory waypoints can either be predicted in Cartesian space (also referred to as task space) or in joint space.
We discuss the pros and cons of both variants and explain why TrueRMA predicts waypoints in Cartesian space.   
With TrueRMA, each waypoint is defined by a fixed number of parameters, meaning that the dimensionality of the action space scales linearly with the number of waypoints and the number of parameters per waypoint. Since the final performance of a trajectory depends on all waypoints, reward can only be given after executing the whole trajectory. 
The reinforcement learning algorithm has to deal with a so-called credit assignment problem \cite{minsky1961steps}.
Reducing the dimensionality of the action vector simplifies the problem, thus making the training phase less data-intensive. 
A waypoint in Cartesian space can be described by three coordinates (x, y, z) and three angles (roll, pitch, yaw).
Dependent on the specific task, the dimensionality might be reducible by setting a fixed value for a specific dimension 
(e.g. a fixed yaw rotation can be used when balancing a ball on a plate).
Moreover, as explained below, the dimensionality can be further reduced by predicting orthogonal adaptations. 
When working in joint space, one dimension per joint is required to define a waypoint.    
The KUKA iiwa robot, which is used for our experiments, has seven degrees of freedom, meaning that seven instead of six dimensions are required to define a waypoint in joint space.   
However, when working in Cartesian space, there are two things to be aware of: 
Firstly, if a robot is kinematically redundant, a pose in Cartesian space can be expressed by multiple joint configurations. 
To avoid generating different trajectories for equal Cartesian waypoints, we use an inverse kinematics solver that returns a unique joint configuration for a given end-effector pose. 
Secondly, it must be ensured that a solution of the inverse kinematics exists for each potential Cartesian waypoint.   
We address this issue by restricting the workspace of the robot accordingly.

\subsection{Recursive Midpoint Adaptations}
\begin{figure}[h]
	\centering
	\vspace{0.2cm} 
	\begin{forest}
		BDT
		[%
		$\hspace{4pt}\mathrm{S}\;\textcolor{black}{\myarrow[0.75pt]}\;\mathrm{E}$
		[$\hspace{10pt}\mathrm{S}\;\textcolor{black}{\myarrow[0.75pt]}\;\mathrm{M_0}$
		[{$\hspace{0pt}\mathrm{S}\;\textcolor{BLUE_ARROW}{\myarrow[0.75pt]}\;\mathrm{M_1}$}, name=l
		]
		[$\hspace{8pt}\mathrm{M_1}\;\textcolor{BLUE_ARROW}{\myarrow[0.75pt]}\;\mathrm{M_0}$, name=m, %
		]
		]
		[$\hspace{-2pt}\mathrm{M_0}\;\textcolor{black}{\myarrow[0.75pt]}\;\mathrm{E}$
		[{$\hspace{0pt}\mathrm{M_0}\;\textcolor{BLUE_ARROW}{\myarrow[0.75pt]}\;\mathrm{M_2}$}, name=r, %
		]
		[{$\hspace{0pt}\mathrm{M_2}\;\textcolor{BLUE_ARROW}{\myarrow[0.75pt]}\;\mathrm{E}$}, %
		]
		]
		]
	\end{forest}
	\caption{Binary tree representation of recursive midpoint adaptations for $N=2$.
		The blue arrows indicate how the waypoints are put together to form a path.\newline
		$S$: starting point; $E$: endpoint; $M_{0..2}$: adapted midpoints }
	\label{fig:binaryTree}
	
\end{figure}
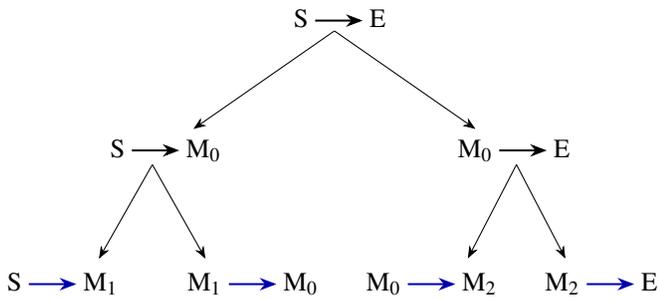

\begin{figure}[h]
	\centering 
	\begin{subfigure}[c]{0.225\textwidth}
		\includegraphics[width=\linewidth]{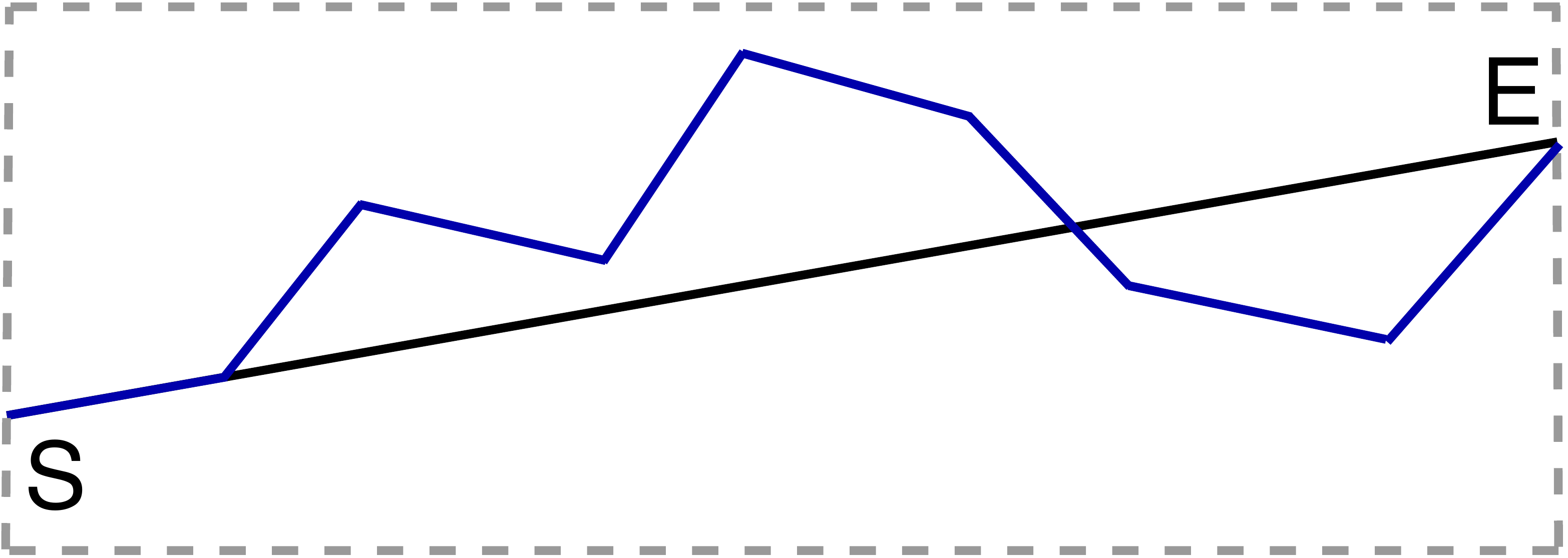}
	\end{subfigure} 
	\begin{subfigure}[c]{0.225\textwidth}
		\includegraphics[width=\linewidth]{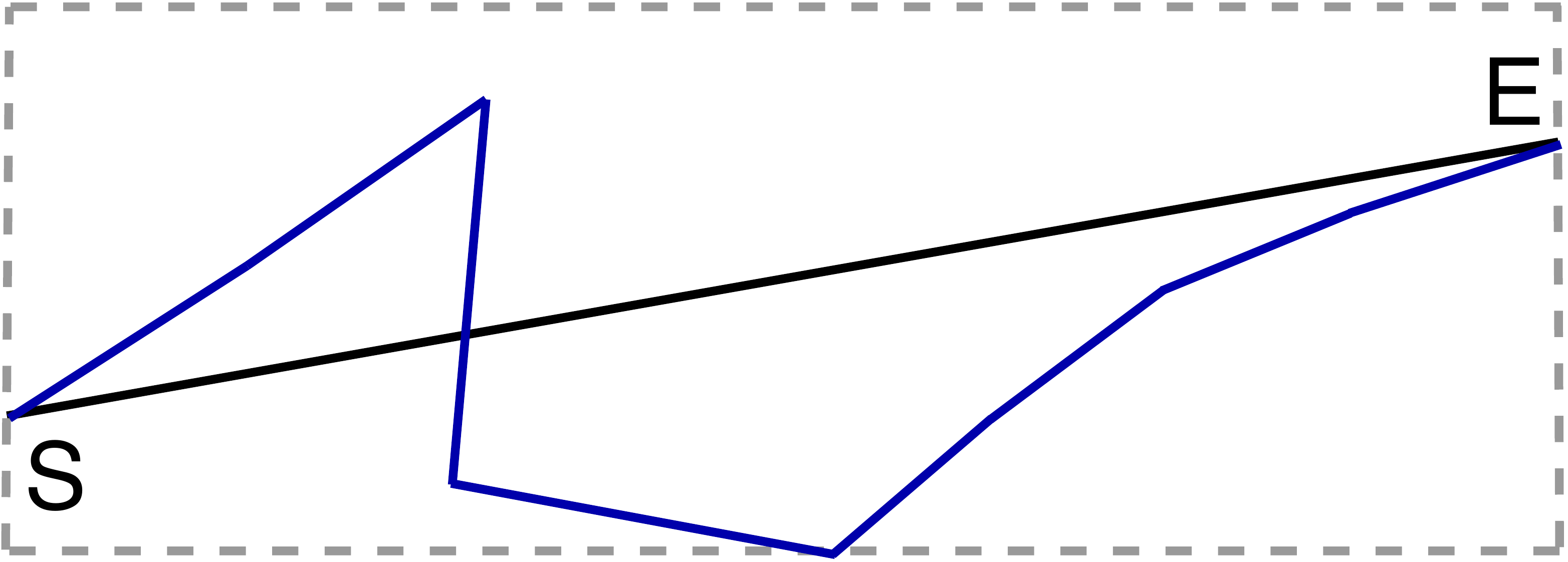}
	\end{subfigure} 
	\caption{Planar paths generated by a random agent with TrueRMA after $N=3$ iterations. The initial straight line is shown in black.}
	\label{fig:varietyRMA}
\end{figure}

An illustrative example for recursive midpoint adaptations is shown in Fig. \ref{fig:RMA}.
The basic idea is to predict adaptations relative to the midpoint of a straight line between a specified starting point and endpoint.
The adapted midpoint splits the path between the starting point and endpoint into two new straight line segments.
By applying the strategy recursively, a path consisting of $2^N +1$ waypoints can be generated within $N$ iterations. 
Fig. \ref{fig:binaryTree} illustrates the recursive approach by means of a binary tree. 
Apart from the leaves, each node represents a single midpoint adaptation. 
In the last step, a new path between the starting point and endpoint is generated by concatenating the adapted midpoints as illustrated by the blue arrows. 
When predicting orthogonal adaptations, the number of parameters per waypoint can be reduced by one. 
While the reduction of the search space simplifies learning, shapes like loops or back-and-forth motions are no longer possible. Fig. \ref{fig:varietyRMA} shows two randomly generated paths for the planar case of orthogonal midpoint adaptations. 

Note that when using orthogonal adaptations, the adapted midpoint is equidistant to both ends of the corresponding straight line. As can be seen in Fig. \ref{fig:varietyBaselines}, the baselines allow arbitrary path shapes.

\subsubsection{The planar case}
In the planar case, one parameter is required to describe a two-dimensional Cartesian point.
The absolute value of the parameter linearly scales the length of the orthogonal adaptation vector, while the sign specifies the direction. In case that the adapted midpoint would lie outside of the specified workspace, the 
length of the adaptation vector is reduced until the adapted point lies on the boundary.
We define the maximum length of the orthogonal adaptation vector to be half of the length of the straight line. 

\subsubsection{The three-dimensional case}
In the three-dimensional case, two parameters specify an adaptation within a plane.
The plane is defined by the unmodified midpoint and the direction of the straight line, which serves as a normal vector.
The parameters are polar coordinates within the plane, meaning that one parameter determines the length of the adaptation vector, while the other specifies an angle in relation to an arbitrarily chosen reference direction. 
The reference direction is chosen in the following way: 
We define a fixed point outside the workspace and calculate $P_{ref}$, the point on the plane with the smallest distance to the fixed point. The line between the unmodified midpoint and $P_{ref}$ defines the reference direction. 

\subsubsection{Orientations}
Each Cartesian orientation (roll, pitch, yaw) is treated as an independent, one-dimensional variable. One parameter is required per orientation.   
The value of the adaptation depends on the parameter and on the length of the corresponding straight line.
Note that adaptations for orientations are one-dimensional.

\subsection{Baselines}
\begin{figure}[h]
	\centering  
	\begin{subfigure}[c]{0.5\textwidth}
		\includegraphics[width=\linewidth]{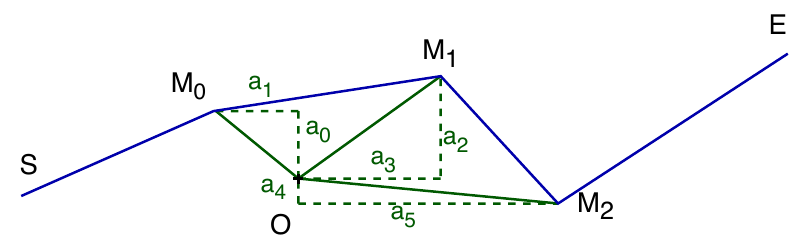}
		\subcaption{TrueAbs}
	\end{subfigure} 
	\begin{subfigure}[c]{0.5\textwidth}
		\includegraphics[width=\linewidth]{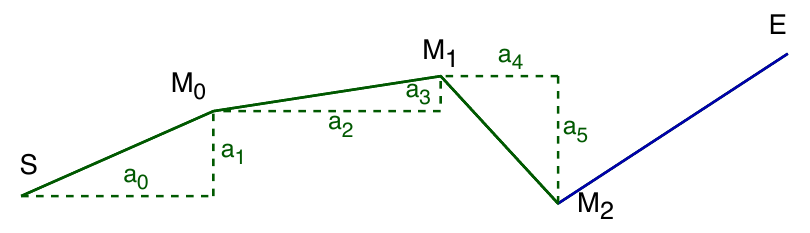}
		\subcaption{TrueRel}
	\end{subfigure}
	\caption{Planar paths with three intermediate waypoints \mbox{specified} by six parameters $a_{0..5}$.}
	\label{fig:waypointsBaselines}
\end{figure}

Two different baselines are implemented in order to assess the performance of TrueRMA.
When using the absolute prediction baseline (``TrueAbs''), actions directly define coordinates of waypoints in Cartesian space.
With the relative prediction baseline (``TrueRel''), actions define adaptations relative to the current Cartesian pose of the end-effector.  
In both cases, the final path is composed of the starting point, the predicted waypoints and the endpoint. Fig. \ref{fig:waypointsBaselines} illustrates the functional principle of the baselines for a planar path with three intermediate waypoints. 
\begin{figure}[h]
	\centering  
	\begin{subfigure}[c]{0.225\textwidth}
		\includegraphics[width=\linewidth]{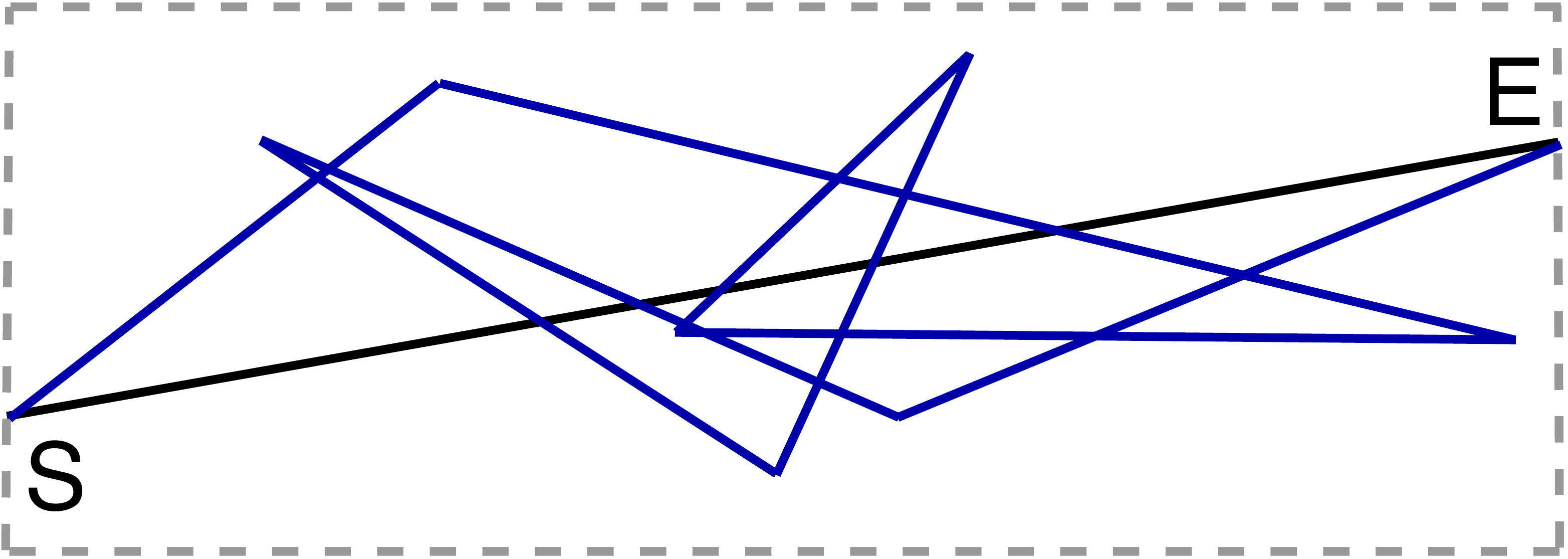}
		\subcaption{TrueAbs}
	\end{subfigure} 
	\begin{subfigure}[c]{0.225\textwidth}
		\includegraphics[width=\linewidth]{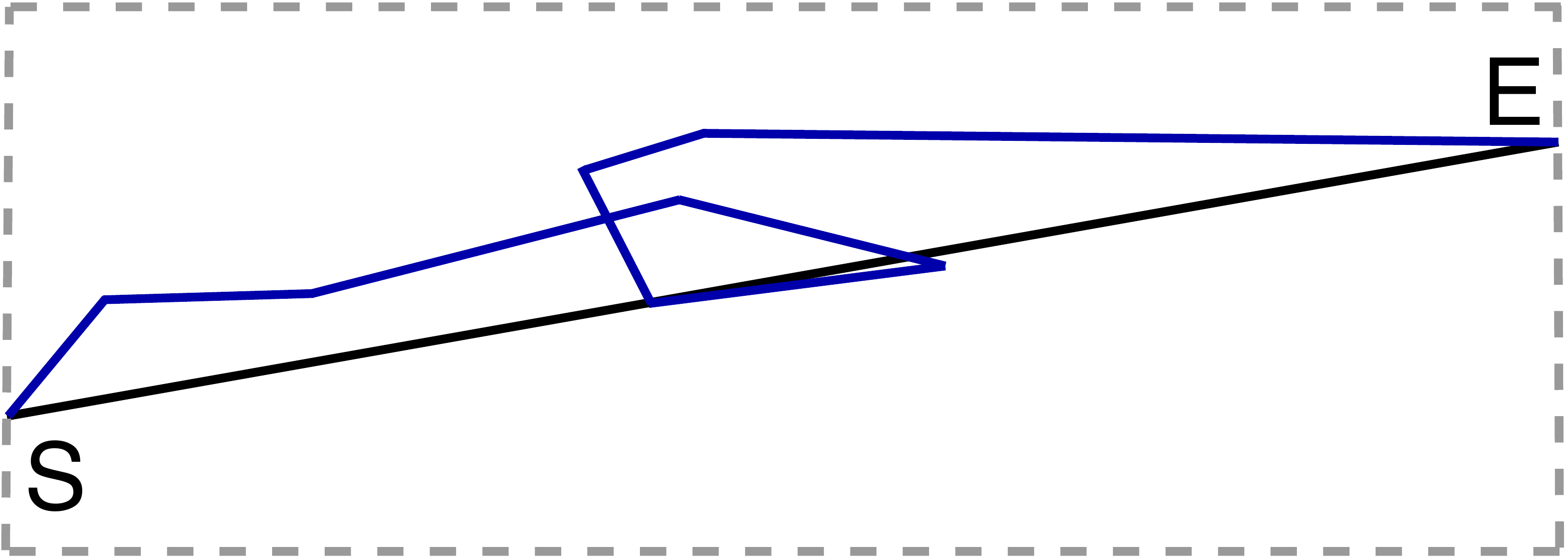}
		\subcaption{TrueRel}
	\end{subfigure}
	\caption{Planar paths with seven intermediate points generated by a random agent.}
	\label{fig:varietyBaselines}
\end{figure}
\vspace{-0.2cm}

\section{Trajectory Generation}
\label{trajectoryGeneration}
\subsection{Smoothing by Adding Circular Blends}

\begin{figure*}[t]
	\centering
	\includegraphics[width=\linewidth, origin=c]{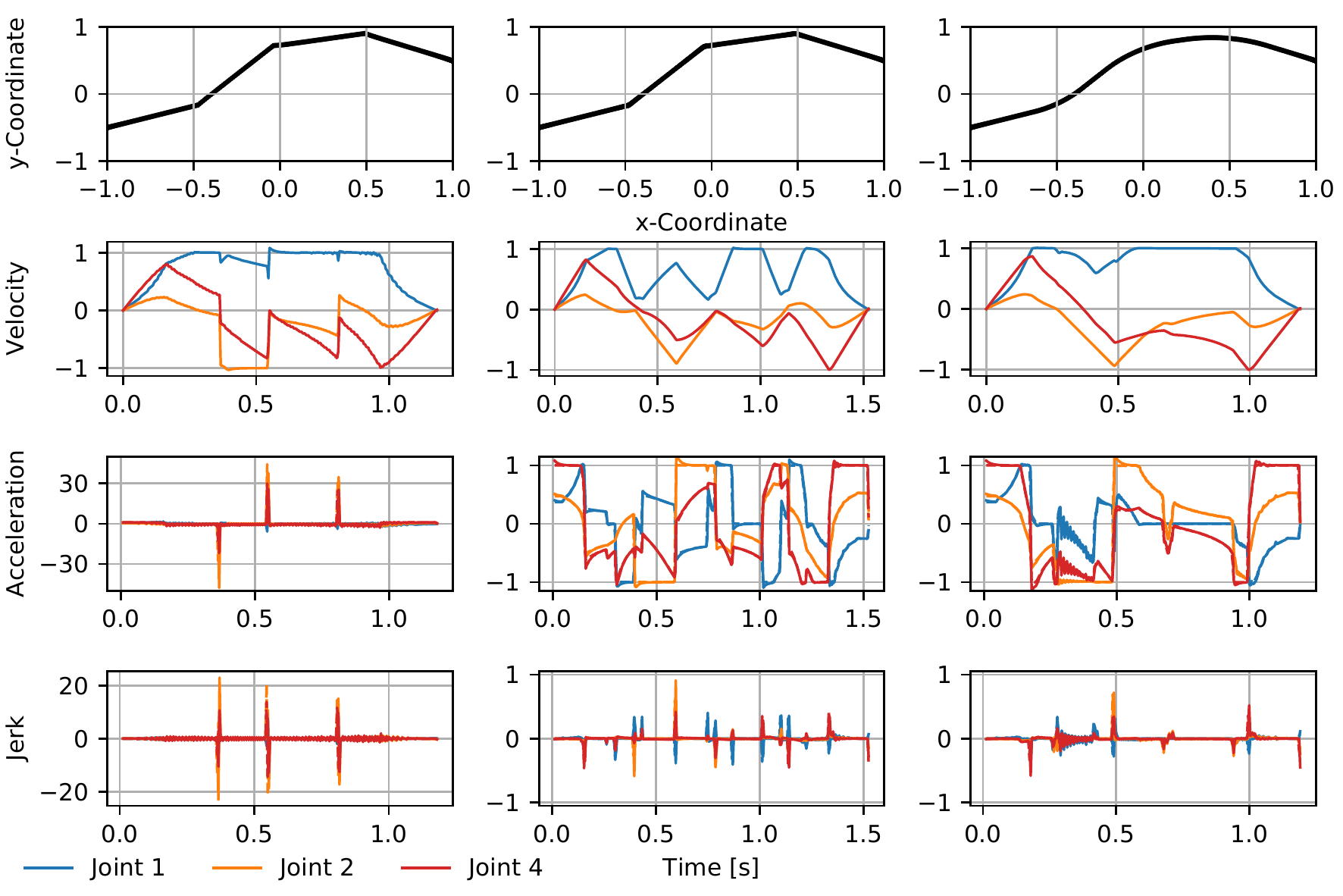}
	\caption{Examples for Cartesian path generation and time-optimal parameterization for the planar case. All metrics are normalized to their boundary values. Missing joints are omitted to preserve a clear overview. \newline
		Left: Without adding circular blends, both acceleration and jerk limits are exceeded.
		Middle: When adding blends with small radii, the parameterization complies with the limits but low joint velocities are required near to the predicted waypoints.  \newline
		Right: Adding blends with greater radii enables fast and smooth trajectory execution.}
	\label{fig:smoothing}
\end{figure*}

Given the Cartesian waypoints from chapter \ref{waypointGeneration}, a differentiable path is generated by connecting circular blends around the waypoints with straight line segments.
The method and its mathematical foundations are described in \cite{kunz2012time}.
While \cite{kunz2012time} performs smoothing in joint space, we add circular blends in Cartesian space. 
Note that due to the circular blends, the smoothed path slightly deviates from the given waypoints.
As shown in Fig. \ref{fig:smoothing}, smoothing is required to avoid high joint accelerations and jerks.

\subsection{Time-optimal Parameterization}
A parameterization is time-optimal with respect to specified limits, if at least one joint is constrained by the limits at each point in time.
The time-optimal parameterization is computed with a method proposed in \cite{kunz2012time}, which supports velocity and acceleration limits.
Although jerk and torque boundaries are not explicitly considered, the corresponding limits were not exceeded in our experiments.
Given a fixed set of waypoints in joint space, a unique parameterization is computed. 
Hence, the neural network does not have to predict the timing of the trajectory. 
While the joint velocity and torque limits are officially provided by KUKA \cite{KUKASpecs}, acceleration and jerk limits are not publicly available. 
Therefore, we resort to the values of a comparable Franka Emika Panda robot \cite{libfranka}.

\section{Evaluation}
\begin{figure}[h]
	\centering
	\begin{subfigure}[c]{0.2255\textwidth}
		\includegraphics[trim=110 4 10 0, clip, width=\linewidth]{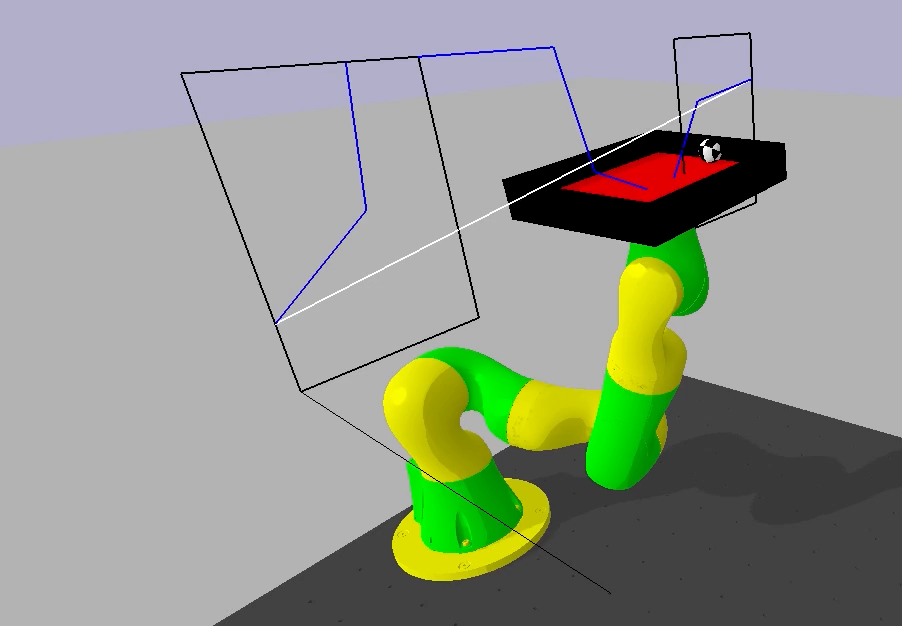}
	\end{subfigure} 
	\begin{subfigure}[c]{0.2475\textwidth}
			\includegraphics[trim=65 20 120 0, clip, width=\linewidth]{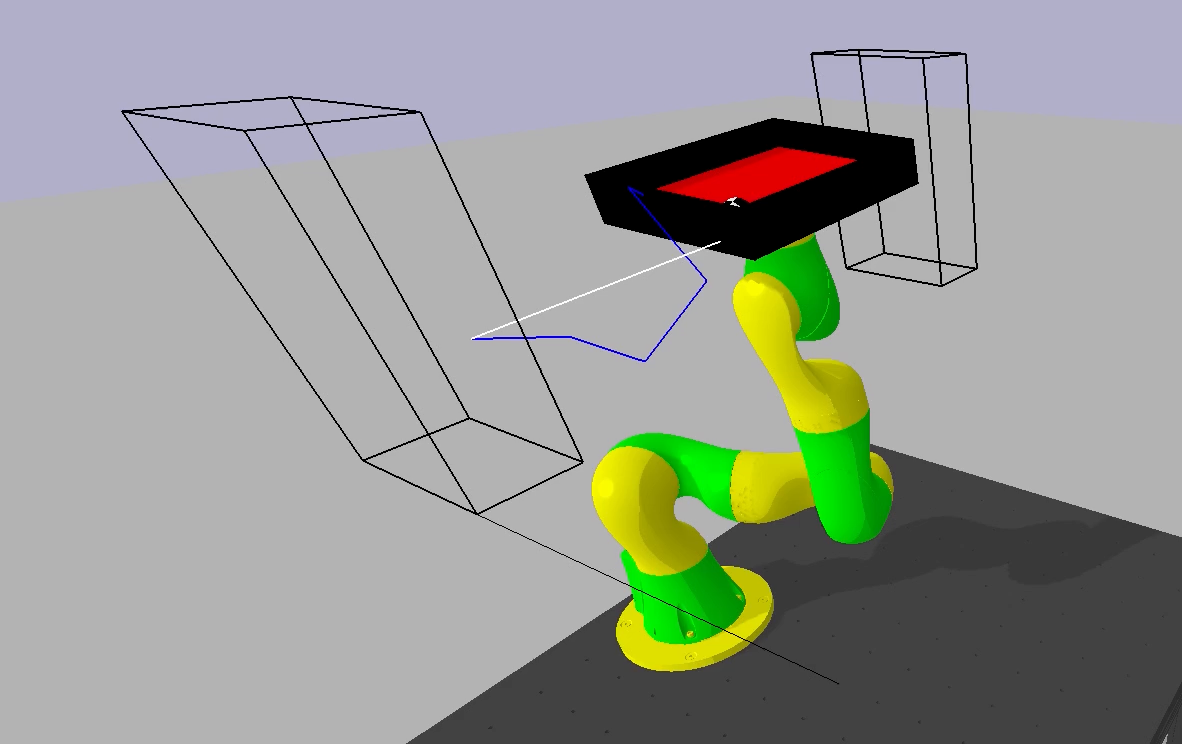}
	\end{subfigure}   
	\caption{Two and three-dimensional ball-on-plate task.}
	\label{fig:ballOnPlate}
\end{figure}

\begin{figure*}[th]
	\centering  
	\vspace{0.2cm} 
	\includegraphics[width=.75\linewidth]{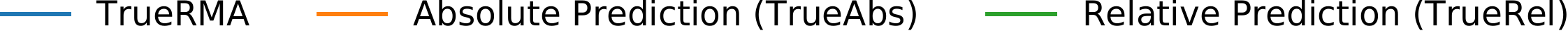}\\
	\begin{subfigure}[c]{0.24\textwidth}
		\includegraphics[width=\linewidth]{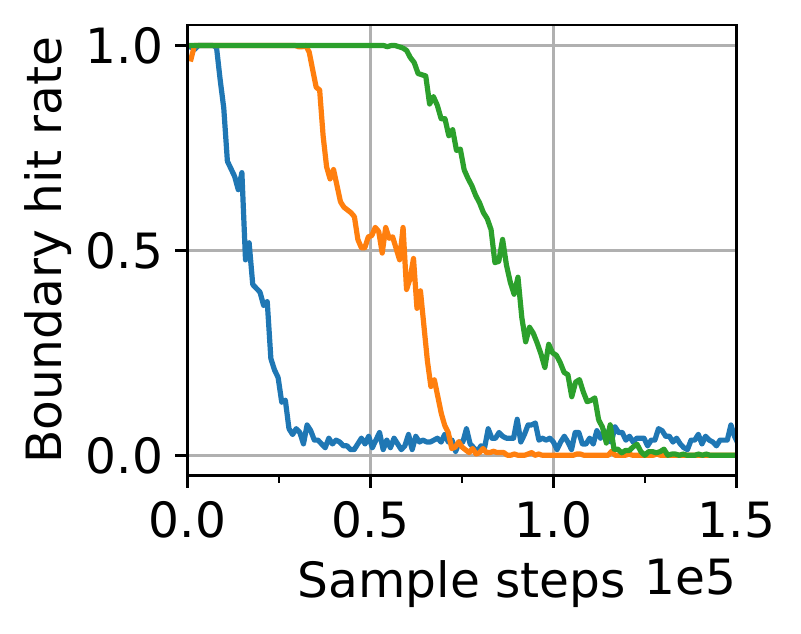}
	\end{subfigure} 
	\begin{subfigure}[c]{0.24\textwidth}
		\includegraphics[width=\linewidth]{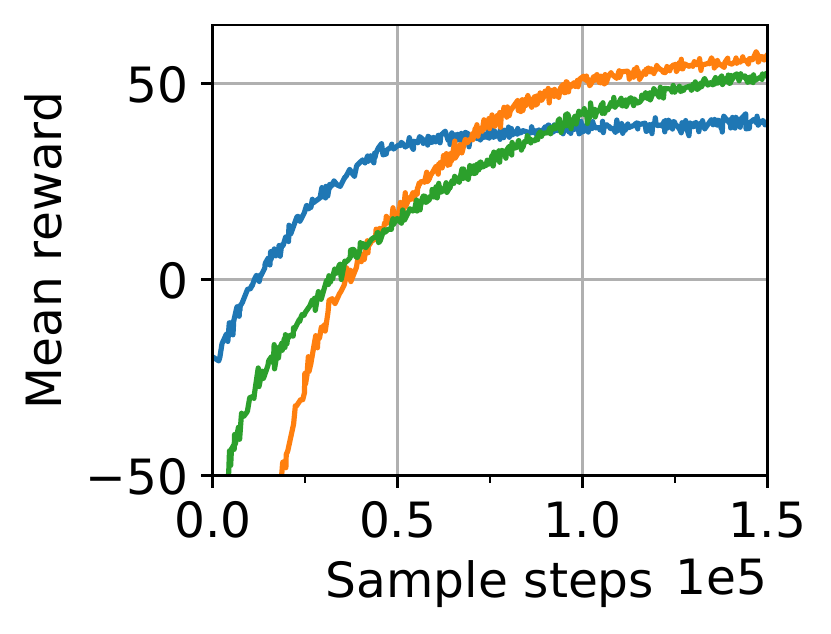}
	\end{subfigure} 
	\begin{subfigure}[c]{0.24\textwidth}
		\includegraphics[width=\linewidth]{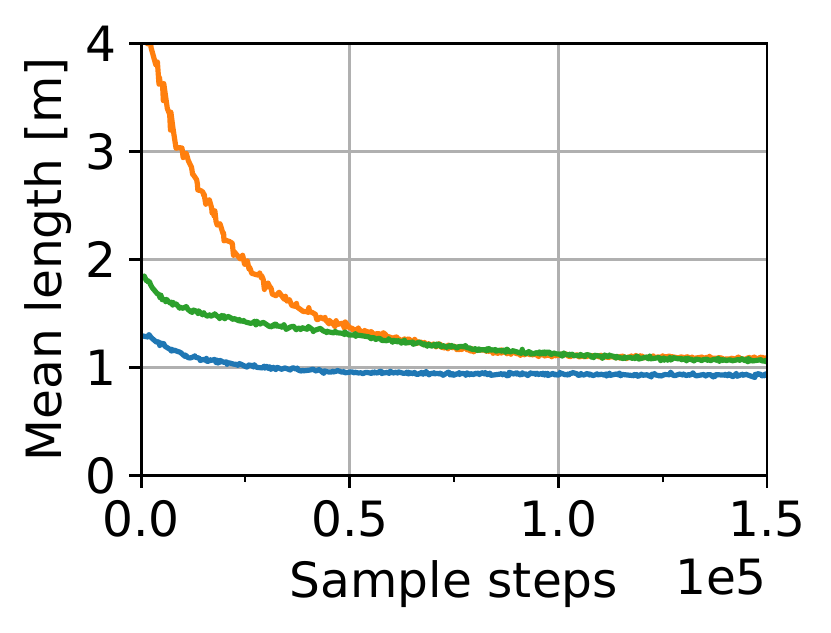}
	\end{subfigure} 
	\begin{subfigure}[c]{0.24\textwidth}
		\includegraphics[width=\linewidth]{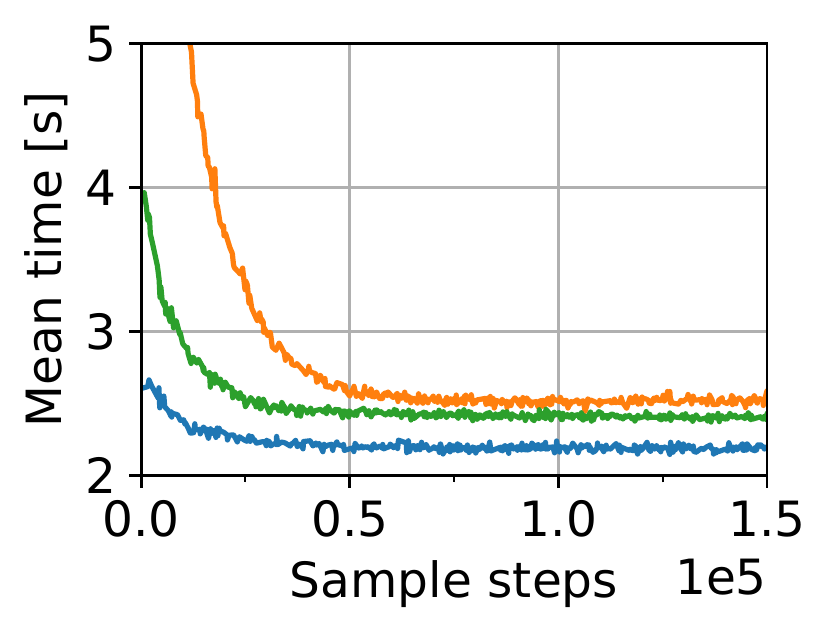}
	\end{subfigure} 
	\caption*{Three-dimensional case}
	\begin{subfigure}[c]{0.24\textwidth}
		\includegraphics[width=\linewidth]{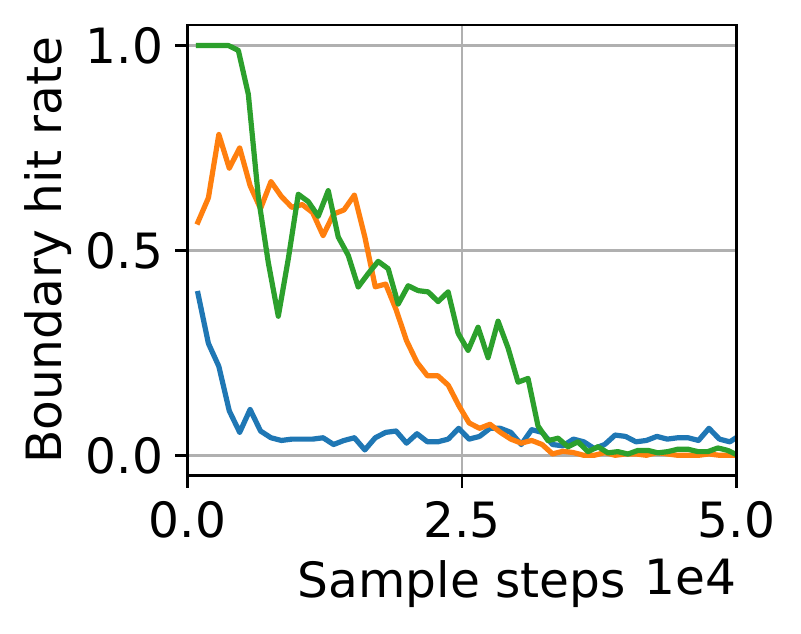}
	\end{subfigure} 
	\begin{subfigure}[c]{0.24\textwidth}
		\includegraphics[width=\linewidth]{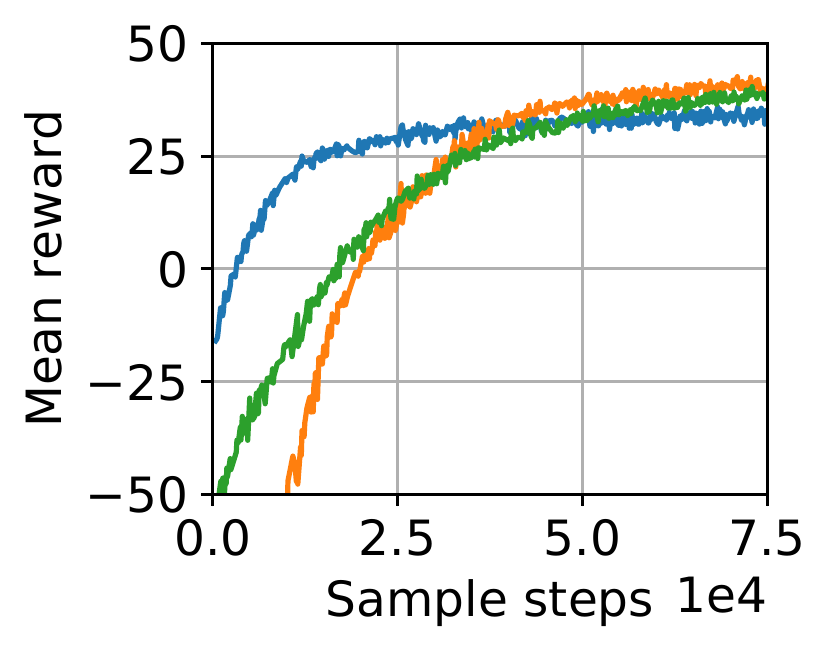}
	\end{subfigure} 
	\begin{subfigure}[c]{0.24\textwidth}
		\includegraphics[width=\linewidth]{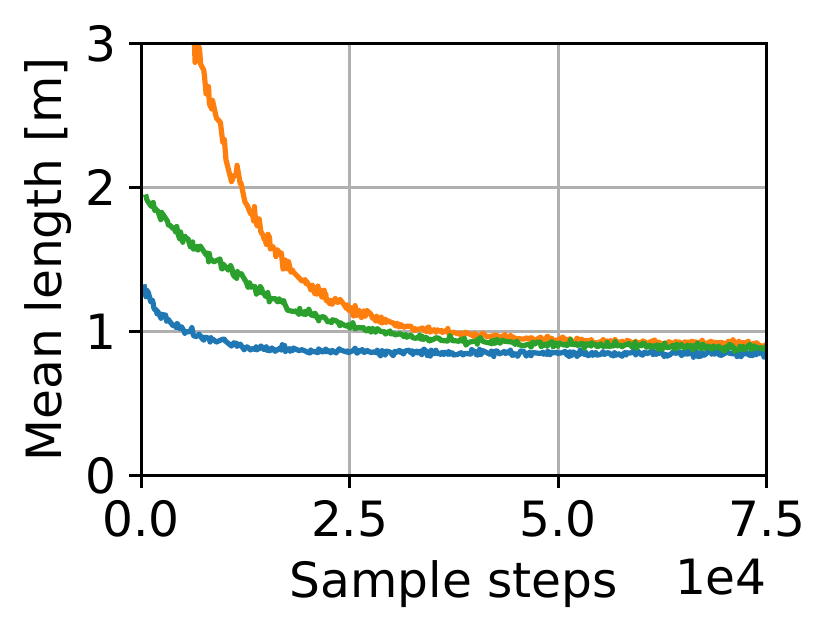}
	\end{subfigure} 
	\begin{subfigure}[c]{0.24\textwidth}
		\includegraphics[width=\linewidth]{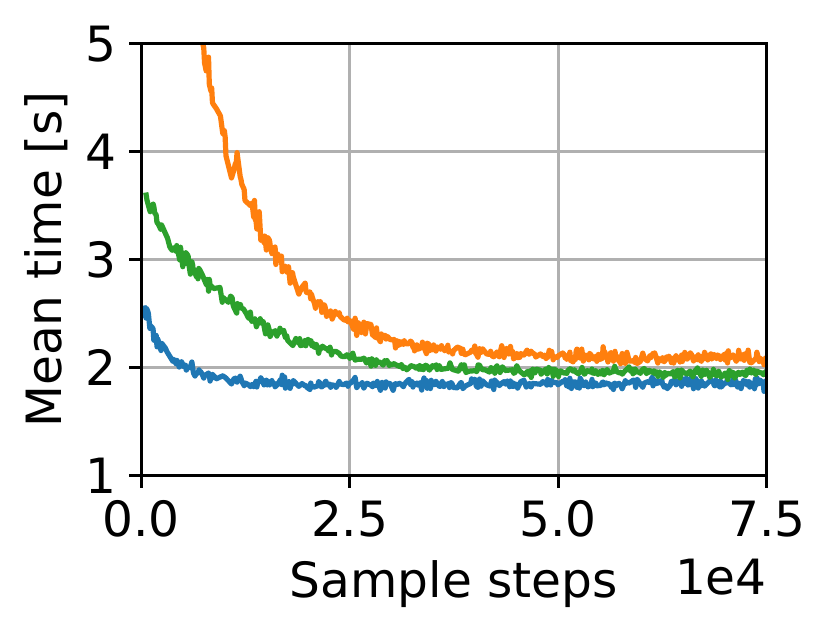}
	\end{subfigure} 
	\caption*{Two-dimensional case}
	\caption{Evaluation criteria of the ball-on-plate task plotted against the progress of the training.}
	\label{fig:results}
\end{figure*}

\subsection{Specification of the Ball-on-Plate Task}
We evaluate our method with a two-dimensional and three-dimensional ball-on-plate task.
The goal is to move between a specified starting point and endpoint without letting a ball touch the boundary of a board. 
Fig. \ref{fig:ballOnPlate} shows the environment for both cases. In the two-dimensional case, the robot moves inside a plane with a size of \mbox{\SI{0.85}{\meter} $\times$ \SI{0.3}{\meter}}. The workspace of the three-dimensional case is a cuboid with dimensions of \SI{0.85}{\meter} $\times$ \SI{0.3}{\meter}  $\times$ \SI{0.2}{\meter}.
The areas edged in black visualize potential starting points and endpoints. During training, the robot is allowed to move from left to right as well as the other way round. Seven intermediate waypoints are predicted.
Further visualization is provided in the accompanying video.  
In the two-dimensional case, pitch angles are adapted, while roll and yaw angles are set to fixed values. Thus, TrueRMA requires two parameters per waypoint. When moving in three-dimensions, the roll angle is adapted as well, leading to four parameters per waypoint. In both cases, the baselines require one additional parameter per waypoint. Reward is assigned according to the cost savings of the adapted trajectory compared to a straight line movement.  The cost is defined as the average distance of the ball from the midpoint of the board multiplied by the execution time of the trajectory.

\subsection{Evaluation Criteria}
The results of the training process for the two and three-dimensional case are shown in Fig. \ref{fig:results} and in the accompanying video. Each sample step corresponds to a single trajectory execution. 
Four different evaluation criteria are selected:

\subsubsection{Boundary hit rate}
The boundary hit rate represents the probability that the ball hits the boundary during trajectory execution.
Since the goal of a ball-on-plate task is to keep the ball on the plate, the boundary hit rate is the clearest indicator 
for the performance of a balancing policy. As can be seen in the left part of Fig. \ref{fig:results}, all methods manage to keep the ball on the plate. However, significantly less data is required when using TrueRMA.
To achieve a hit rate of less than \SI{10}{\percent} in the three-dimensional case, TrueRMA required 27 000 trajectory execution, while the absolute prediction baseline and the relative prediction baseline required 69 000 and 112 000 trajectory executions, respectively. 

\subsubsection{Mean reward}
The reward is specified as the performance gain compared to a straight line movement. A reward of zero means that the generated trajectory performs as well as a straight line movement. 
Fig. \ref{fig:results} shows that the performance of TrueRMA rises significantly faster than the performance of the baselines.   
However, TrueRMA converges at a lower final reward, meaning that the baselines perform better if sufficient data is available. The results are plausible, since the baselines can learn arbitrary path shapes, while TrueRMA restricts the search space to accelerate learning.

\subsubsection{Mean trajectory length and execution time}
Compared to the baselines, TrueRMA generates the shortest trajectories. 
This outcome is reasonable as the search is initialized from a straight line, which is the shortest possible path. 
As can be seen in Fig. \ref{fig:varietyBaselines}, the absolute prediction baseline starts with very long trajectories. 
However, after training for a sufficiently long time, relatively short trajectories are generated. The trajectories generated by TrueRMA are the fastest, which is plausible as they are shorter. We note that the learning process aims to optimize both balancing performance and execution time, which means that faster trajectories do not necessarily lead to a higher reward.

\section{CONCLUSIONS}

We presented a workflow to learn smooth and fast robot trajectories in Cartesian space and evaluated three different methods to generate waypoints based on network predictions.  All three methods succeeded in keeping a ball on a plate while moving between specified points. However, significantly less training data was required when predicting orthogonal midpoint adaptations.

In future work, we intend to evaluate the performance of our approach for learning collision-free trajectories in the presence of moving obstacles.

\section*{ACKNOWLEDGMENT}

This research was supported by the German Federal Ministry of Education and Research (BMBF) and the Indo-German Science \& Technology Centre (IGSTC) as part of the project TransLearn (01DQ19007A).

\bibliographystyle{IEEEtran}
\bibliography{root}

\end{document}